\documentclass{article}

    \usepackage[preprint, nonatbib]{neurips_2020}

\usepackage[utf8]{inputenc} % allow utf-8 input
\usepackage[T1]{fontenc}    % use 8-bit T1 fonts
\usepackage{hyperref}       % hyperlinks
\usepackage{url}            % simple URL typesetting
\usepackage{booktabs}       % professional-quality tables
\usepackage{amsfonts}       % blackboard math symbols
\usepackage{nicefrac}       % compact symbols for 1/2, etc.
\usepackage{microtype}      % microtypography

\usepackage[pdftex]{graphicx}
\usepackage{amsmath}
\usepackage{booktabs}

\usepackage{subfig}
\usepackage{graphicx}

\title{Prune Responsibly}

\author{Michela Paganini \\
Facebook AI Research \\
\texttt{michela@fb.com} \\
}

\begin{document}

\maketitle

\begin{abstract}
    Irrespective of the specific definition of fairness in a machine learning application, pruning the underlying model affects it.
    We investigate and document the emergence and exacerbation of undesirable per-class performance imbalances, across tasks and architectures, for almost one million categories considered across over 100K image classification models that undergo a pruning process. 
    We demonstrate the need for transparent reporting, inclusive of bias, fairness, and inclusion metrics, in real-life engineering decision-making around neural network pruning.
    In response to the calls for quantitative evaluation of AI models to be population-aware, we present neural network pruning as a tangible application domain where the ways in which accuracy-efficiency trade-offs disproportionately affect underrepresented or outlier groups have historically been overlooked. We provide a simple, Pareto-based framework to insert fairness considerations into value-based operating point selection processes, and to re-evaluate pruning technique choices.
    
\end{abstract}
\section{Introduction}
Large-scale deep learning models are presently being employed in society at large across a multitude of high-stakes applications, including healthcare, law enforcement, and insurance, with consequential decision-making power over various aspects of human life. In order to reduce the memory and computational footprints, as well as the energy consumption of these models, techniques for model compression, including pruning and quantization, are often used to alter the model in invasive yet supposedly uninfluential ways.

Thresholds are usually placed on acceptable performance loss in compressed models. However, it is currently uncommon, when introducing new pruning techniques, to benchmark them according to fairness and bias metrics, and to further investigate the downstream effects of even minimal performance degradations introduced by the proposed intervention. Our goal is to promote proactive monitoring, especially among the community involved in real world deployment of pruned models, of the subtle dangers and repercussions of applying model compression interventions without a proper bias and fairness evaluation pipeline. 

Systematic biases are known to plague many AI models~\cite{buolamwini2018gender}, and may be exacerbated in non-trivial ways by pruning interventions.
We are interested in quantifying the inequality of treatment among classes, cohorts, and individuals as network capacity is reduced. Specifically, the hypothesis we seek to test is that class imbalance and class complexity affect the per-class performance of pruned models. We model and measure the contribution of these factors to the observed changes in performance in pruned models, while controlling for other factors of variations, such as model type, dataset, pruning technique, and choice of rewinding or finetuning weights after pruning. We demonstrate that, as networks are pruned and sparsity increases, model performance deteriorates more substantially on underrepresented and complex classes than it does on others, thus exacerbating pre-existing disparities among classes. In addition, as model capacity is reduced, networks progressively lose the ability to remain invariant to example properties such as angular orientation in images, which results in higher performance losses for groups, sub-groups, and individuals whose examples exhibit these properties.

We critique the practice of benchmarking the efficacy of novel pruning routines solely along axes of total accuracy and computing costs, and suggest fairness as an additional dimension to consider when evaluating trade-offs and disclosing limitations. In line with prior work on ethical reporting~\cite{model-cards, raji2019ml}, we raise awareness on the problematic reporting norms in the model compression literature.

As recommended~\cite{raji2019ml}, we refrain from proposing yet another complex and opaque technical solution to the socio-technical problems of unfairness and bias in AI models; instead, we limit ourselves to documenting and quantifying the extent of these issues in a common, practical scenario, while contributing methodology for operationalizing this type of assessment in ways that can be integrated into transparent decision-making workflows for practitioners. We believe this to be of particular urgency for model compression and architecture optimization, due to the growing practical need of reducing model size, especially in applications of AI in developing countries, on device, and in low-bandwidth or otherwise low-resource environments~\cite{li2016pruning, paganini2020streamlining}.

\section{Related Work}
Prior work on machine learning transparency recommended the release, alongside new models, of detailed and transparent performance documentation and intended use-cases in an Acceptable Use Policy format called \textit{model cards}, in order to prevent model misuse in unsafe scenarios~\cite{model-cards}. Similarly, our work calls for more thoughtful and responsible practices around evaluation and transparent reporting of model performance, all while specifically grounding the analysis in the concrete use-case scenario of neural network pruning, as demanded by work that advocates for moving beyond abstract and unrealistic fairness settings~\cite{doi:10.1177/2053951717743530}.

Surveys of ML practitioners across companies and roles, aimed at better understanding the practical challenges they face when trying to build fairer systems in real-world contexts~\cite{Holstein_2019}, served as an inspiration for us to focus on tackling challenges in a concrete scenario that often occurs in model design and optimization across ML products: pruning. We identify pruning as a high-stakes example of a widespread yet potentially disruptive technique for model compression used in real-world, deployed models. Specifically, in this work, we demonstrate the need for transparent reporting, inclusive of bias, fairness, and inclusion metrics, as advocated for in prior related work~\cite{model-cards}, in real-life engineering decision-making in the context of pruning. 

Others have pointed out how the very process of documenting details of model design and expected behavior induces practitioners to reflect on the practical consequences of their engineering choices throughout the whole model design and development procedure~\cite{raji2019ml}.
In this work, we apply notions and recommendations from prior foundational ML fairness and transparency literature to demonstrate its applicability as an anchor for ethical AI engineering and development. Concretely, we respond to calls for transparent reporting and accounting of sub-population and intersectionality in model performance, by presenting a practical proving ground, in the form of the network pruning scenario, in which traditional trade-off considerations often fail to examine effects on individuals, populations, and classes. 

Classical trade-offs between accuracy and fairness may be unjustifiable or unethical, represent a practical impossibility in many high-stakes scenarios, or simply create false dichotomies~\cite{chen2018classifier}. We suggest, instead, considering sparsity-fairness trade-offs when neural network pruning is applied, and present these two quantities under the light of two desirable yet, at times, competing secondary goods.

Finally, issues of compatibility of fairness quantification strategies with one another or with the legal framework~\cite{friedler2016impossibility, verma2018fairness, corbettdavies2018measure, mitchell2018predictionbased, xiang2019legal} guided our decision to abstain from proposing new context-specific fairness measures or automated fairness-inducing or unfairness-mitigating techniques.

\section{Experiments}
\label{sec:experiment}
We make two levels of arguments as to why those deploying pruned models should consider fairness as part of their selection process. First, we consider a network population argument, to assess the effect of class characteristics and imbalance on per-class pruned network performance, as quantified from the analysis of almost one million classes considered across over 100K trained and pruned image classification models. Second, we consider two individual case studies to highlight practical implications of fairness evaluation in pruned models; in one, we are able to peek behind latent and sensitive attributes, such as group membership, to gain some understanding of reasons behind degraded pruned network performance targeting certain individuals (Sec.~\ref{sec:qmnist-study}); in the other, we borrow from the tools of economics to provide a framework and real-world example of how including fairness may affect pruning method selection (Sec.~\ref{sec:pareto}).

To ascertain the global effect of pruning on class-level accuracy, we conduct a large-scale experimental study, training a series of LeNet~\cite{LeCun1990-em}, AlexNet~\cite{alexnet}, VGG11~\cite{vgg} and ResNet18~\cite{he2015deep} architectures in a supervised way on the following image classification tasks: MNIST~\cite{mnist:1994}, Fashion-MNIST~\cite{fashionmnist}, QMNIST~\cite{yadav2019cold}, Kuzushiji-MNIST (or KMNIST)~\cite{clanuwat2018deep}, EMNIST~\cite{cohen2017emnist}, CIFAR-10, CIFAR-100~\cite{krizhevsky2009learning}, SVHN~\cite{netzer2011reading}.
Each iteration of training consists of 30 epochs of stochastic gradient descent with constant learning rate 0.01 and batch size of 32, across 4 NVIDIA Tesla V100-SXM2 GPUs using \texttt{torch.nn.DataParallel}~\cite{NEURIPS2019_9015} to handle distribution.
Each network is iteratively trained, pruned, and either finetuned or rewound to the original weights at initialization (as suggested in the Lottery Ticket-finding algorithm~\cite{frankle2018lottery}), for 20 iterations. Each pruning iteration was set to remove 20\% of the unpruned weights. We repeat the procedure for $L_1$, $L_2$, $L_\infty$, and random structured pruning, and $L_1$, global, and random unstructured pruning. More details on the pruning techniques can be found in Appendix~\ref{appendix:pruning}. All together, this consists of 120,906 models for analysis.

We evaluate all models at a per-class level, in order to measure per-group performance degradation of pruned models, with class-accuracy for a given class $c$ defined as

\begin{equation}
a_c = \frac{|\{ (x, y) \in \mathcal{D} \  |\  y = \hat{h}(x) = c \}|}{|\{(x, y) \in \mathcal{D}\ | \ y(x) = c \}|},
\end{equation}

where $\mathcal{D} = \{(x_i, y_i)\}_{i}$ is a labeled classification dataset where $x_i\in\mathbb{R}^D$ and $y_i\in\mathbb{N}$, and $\hat{h} : \mathbb{R}^D \longrightarrow \mathbb{N}$ is a trained model we wish to evaluate.
As networks are iteratively sparsified, total accuracy drops, but it does so at non-identical rates across classes (Fig.~\ref{fig:alexnet_mnist_fair}; see Figs.~\ref{fig:perclass}, and~\ref{fig:qmnist_acc_perclass_GU_LeNet} in the Appendix for more detail).

\subsection{Model}

For a pruned model, we wish to assess the impact of variables of interest, such as class imbalance or class complexity, on the per-class performance, to assess disparate class treatment in pruned models. We hypothesize that even minor performance drops observed (and often deemed acceptable) in pruned models might be disproportionately affecting the model's performance on individual classes, as opposed to being evenly distributed among all classes. 
We model the class-accuracy as a function of all measured covariates using an ordinary least squares (OLS) fit to the collected pruned model performance data. 

For simplicity, we begin by only including linear terms in the explanatory variables, and later compare to a second model that adds second order interaction terms between them. Note that, in addition to the polynomial terms, the relationship between accuracy and model sparsity is further modeled through an additional exponential term -- a reasonable modeling assumption supported by prior knowledge of accuracy-sparsity curves in the pruning literature~\cite{frankle2018lottery, Gale2019-xx, morcos2019ticket, zhou2019deconstructing, renda2020comparing, blalock2020state, frankle2020early}.

Our model
\begin{align}
\label{eq:model_formula}
\begin{split}
    a_c = &\ A\cdot e^{\mathrm{sparsity}} 
    + B\cdot \mathrm{sparsity}
    + C\cdot a^0_c 
    + D\cdot \mathrm{quartile}(a^0_c)
    + E\cdot{r_c} + \\
    & + F\cdot{H_c}
    + \bf{G}\cdot I_d
    + \bf{H}\cdot I_w 
    + \bf{J}\cdot I_p 
    + \bf{K}\cdot I_m 
    + \mathrm{cross \ terms}
\end{split}
\end{align}
accounts for the following:

\begin{enumerate}
    \item \textbf{unpruned model accuracy}, $a^0_c$, quantified as the initial, pre-pruning accuracy for class $c$;
    \item \textbf{sparsity}, defined as the fraction of pruned weights;
    \item \textbf{class imbalance}, $r_c$, quantified in terms of the over- or under-representation quantity $r_c = n_c - \frac{1}{C} \displaystyle\sum_{i=1}^C n_i$,
where $n_i = |\{(x, y) \in \mathcal{D}\ | \ y = i \}|/|\mathcal{D}| $ is the fraction of examples belonging to class $i$ that are present in the training set, and $C$ is the number of classes in that classification task;
    % \item \textbf{intra-class entropy}, $v_c = \frac{1}{P} \displaystyle\sum_{k=1}^P \left[-\displaystyle\sum_{w}(p_w \ \log_2(p_w))\right]$, defined as the per-pixel entropy across all images in class $c$ averaged over all $P$ pixels in an image, where the inner sum is taken over all pixel values $w$ for pixel $k$, with $p_w$ being the number of images in the dataset for which pixel $k$ has value $w$;
    \item \textbf{average example entropy}, $H_c = \frac{1}{C} \displaystyle\sum_{i=1}^{N_c} \left[-\displaystyle\sum_v(p_v \ \log_2(p_v))\right]$, defined as the average Shannon entropy of images in class $c$, where the inner sum is taken over pixel values $v$, with $p_v$ being the number of pixels in the image that have value $v$;
    \item \textbf{categorical variables}, $I_m$, $I_d$, $I_w$, and $I_p$, that indicate, respectively, the model, dataset, weight treatment, and pruning technique.
\end{enumerate}

Including the unpruned network's accuracy, $a^0_c$, in our statistical model allows us to separately account for class performance disparities that we know to be generally present in AI models, irrespective of pruning. We aim to assess what causes them to get exacerbated during pruning. This modeling choice disentangles that effect from those of other factors, while explicitly accounting for it as a predictor of post-pruning performance.

Though not all variables are direct quantities of interest, such elements of our design matrix ensure we are controlling for factors of experimental variability. In other contexts, wherever relevant, the model could include sensitive group belonging indicators and group descriptors, as well as higher order interaction terms to address intersectionality, in order to quantify how performance varies across sub-populations.

From the fit (see Appendix~\ref{appendix:fit} for fit diagnostics), we observe statistically significant evidence that both class imbalance and class complexity (as measured by $r_c$ and $H_c$) affect per-class performance by yielding lower class accuracy for underrepresented and complex classes, when controlling for other experimental factors. This implies that underrepresented and more complex classes are most severely affected by pruning procedures.
This is line with well-documented findings of disproportionate misclassification rates for underrepresented gender, skin-type, age, nationality, income, and intersectional demographic groups in computer vision applications~\cite{phillips2003face, grother2011report, 6327355, buolamwini2018gender, devries2019does}, but adds the dimension of network-modifying actions, such as pruning, to the conversation. 

While one of the primary causes of performance degradation in pruned models across classes is, of course, model sparsity itself, after controlling for this variable, at all levels of sparsity, as sparsity increases, not all classes exhibit identical levels of accuracy loss, with the difference being driven, among others, by class imbalance and complexity.

The fit achieves an adjusted $R^2$ of 0.753.
With $\alpha\ll0.01$ (one-tailed) significance, we reject the null hypotheses of $E = F = 0$ in Eq.~\ref{eq:model_formula} being zero, in favor of the alternative hypotheses of $E > 0$ and $F < 0$ respectively. As $r_c$ decreases, the class becomes less represented. Then, $E > 0$ implies that the expected final accuracy post-pruning is lower when class representation is lower. $F < 0$ implies that the expected final accuracy post-pruning is lower for higher entropy classes.

\subsection{Limitations}
Since the response variable -- in this case, accuracy -- is bounded between 0 and 1, the assumptions of OLS do not fully hold. A generalized linear model with an appropriate link function could be adopted to further improve the fit, at the cost of model simplicity.

Due to the high computing costs incurred in such a large scale study, we do not experiment with, nor are we able to assess the impact of various choices around the training strategy, such as changes in the optimization algorithm, or in optimization hyper-parameters such as learning rate, batch size, or number of iterations. Including such terms would allow to further constrain the contribution of variables of interest like imbalance in class representation in the training set.

\section{Case Studies}
After empirically demonstrating the impact of class attributes and representation on class performance in pruned models across a generic set of image classification tasks, models, and pruning techniques, we now present narrow, practical case studies that surface specific challenges that emerge in designing and employing pruned models. 

\subsection{Pruned QMNIST Model}
\label{sec:qmnist-study}

\begin{figure}
    \centering
    \includegraphics[width=0.4\textwidth]{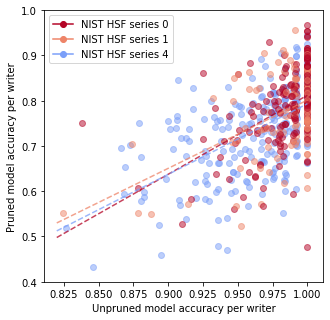}
    \caption{Average accuracy of the pruned and unpruned model on the digits produced by every writer in the test set. Each point represents an individual, color-coded by writer group. The lines represent the best linear fits through the data from the corresponding group.}
    \label{fig:qmnist_acc_old_new}
\end{figure}

\begin{figure}
    \centering
    \includegraphics[width=0.325\textwidth]{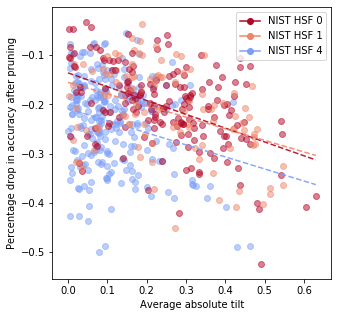}
    \includegraphics[width=0.325\textwidth]{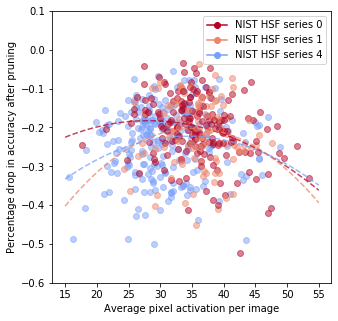}
    \includegraphics[width=0.325\textwidth]{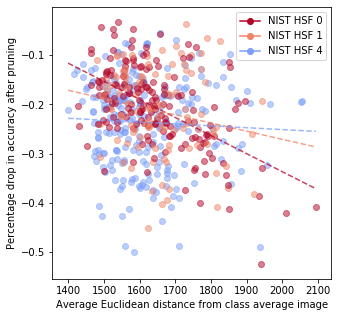}
    \caption{Effect of image properties on model performance per writer.}
    \label{fig:qmnist_percaccdrop_averages}
\end{figure}

In a feat of computational archaeology, the QMNIST dataset~\cite{yadav2019cold} extends the classical MNIST dataset~\cite{mnist:1994} by recovering additional test images from the original NIST set~\cite{grother1995nist} from which MNIST was derived, along with useful metadata such as each digit writers' ID and group. This information can be used to monitor the performance of a model and its pruned version on groups of writers and individual writers. In particular, we will show how pruning alters model performance in non-trivial ways, with accuracy drops disproportionately affecting certain individuals along group or handwriting style lines.

The digits contained in this dataset were handwritten by different groups of writers: the series identified as \textit{hsf4} was produced by high school students and is known for being more challenging, while series \textit{hsf0} and \textit{hsf1} were produced by NIST employees~\cite{mnist:1994, repoqmnist}. In the test set, each individual contributed between 34 and 134 digits.

We train a LeNet architecture (see Appendix~\ref{app:qmnist} for experiment details) to perform digit classification on the QMNIST training set, then iteratively prune and retrain the model for 20 lottery-ticket-style iterations~\cite{frankle2018lottery} using global unstructured pruning, to achieve a total sparsity of ${\sim}{98.6\%}$, corresponding to approximately 800 parameters, which, incidentally, is comparable to the dimensions of the $28\times28$ input space of the task. The total accuracy on the test set degrades from an average of ${\sim}{97\%}$ to an average ${\sim}{78\%}$ due to pruning, across six experimental replicates.

The per-individual performance for the pruned and unpruned models are displayed, for one of the experiments (seed 0), in Fig.~\ref{fig:qmnist_acc_old_new}, along with group annotations. For all three groups of authors, the performance degradation follows similar trends, suggesting an apparent impartiality of the pruning process towards group membership. However, the model's accuracy on the digits produced by certain individuals is significantly reduced after pruning, as is the case for writer 469, on whose digits the model's accuracy decreases from 100\% to 47.7\%.

We investigate the nature of the adverse effects that pruning is exacerbating in this model, and shed light on some of the high-level concepts that are forgotten upon parameter removal. We notice, for example, a negative relationship between average digit tilt and performance change post-pruning (Fig.~\ref{fig:qmnist_percaccdrop_averages}). For further details on the impact of tilt on model performance before and after pruning, see Appendix~\ref{app:qmnist}. Similarly, the average pixel activation (which is itself correlated with digit thickness, elaborateness, and knottiness) correlates with larger drops in performance after pruning for writers that deviate from the mean in either direction. 
On the other hand, average Euclidean distance of each writer's digits to the mean training image in the corresponding class shows different trends with respect to its effect on accuracy drop, depending on the the writer's group membership (Fig.~\ref{fig:qmnist_percaccdrop_averages}). This surfaces a difference in the model's treatment of different demographics of writers, which is important in correctly explaining the network's behavior in a transparent manner. However, it wouldn't have been possible to uncover this without the rich metadata of the QMNIST dataset, which underscores the need for more academic datasets with demographic annotations or, in the context of ML systems, for additional feature and data collection for sub-population identification.

Furthermore, Fig.~\ref{fig:qmnist_percaccdrop_averages} points to the network's ability to conditionally make use of internal representations related to these high-level concepts, to aid in the classification of examples that require knowledge of those quantities; it also points to the problems that might arise when the model's capacity is reduced (for example, through pruning) and the net is no longer being able to compute and use said representations, or related invariances. This will more noticeably affect the cohorts of writers for which those quantities are discriminative and useful towards digit classification, thus making the act of pruning not group independent. 

Although the group performance, in terms of average accuracy drop, is comparable across the three groups, disparities in group and individual treatment have emerged upon further analysis of the pruned network's behavior, signaling a need for caution and heightened attention towards group and individual fairness metrics when deploying pruned models. 
\subsection{Fairness-Aware Decision-Making in Neural Network Pruning with Pareto Optimality}
\label{sec:pareto}
Starting from the finding that class accuracy varies as a function of factors like class representation, average class entropy, intraclass variability, and class complexity in general, we show, borrowing from the tools of economics, one way to include group fairness considerations in the selection of pruning technique to adopt in a realistic application scenario. 

Let's assume we have a large, over-parametrized classifier to productionalize. We can take the AlexNet architecture trained on MNIST as a proxy for this regime. To take full advantage of our hardware, the task is to select the best structured pruning technique among the ones experimented with. However, in real-world production systems, only minimal performance losses are tolerated, in exchange for memory and inference savings. To simulate this constraint, we can enforce that the accuracy not fall below 98\%. Our suggested course of action is the following: instead of simply selecting the pruning technique that yields the sparsified network with the largest sparsity without violating the accuracy requirement, insert fairness as an additional \textit{good} to consider, and select an operating point based on the explicit value assigned to each good.
To do so, we advocate taking into consideration the Pareto-optimality of each pruning technique with respect to sparsity and an application-specific definition of fairness. In this case, a pruning technique is Pareto-optimal if there is no other candidate pruning technique that is weakly both fairer and sparser. In such a setting, an engineer working to select a pruning operating point ought to select a method on the Pareto frontier, the set of all Pareto optimal points.

\begin{figure}
\centering
   \subfloat[ Per-class accuracy breakdown as a function of sparsity for $L_1$ structured pruning with weight rewinding and retraining between each iteration. Each class is color-coded according to the unpruned network’s class performance.]{\includegraphics[width=0.31\textwidth]{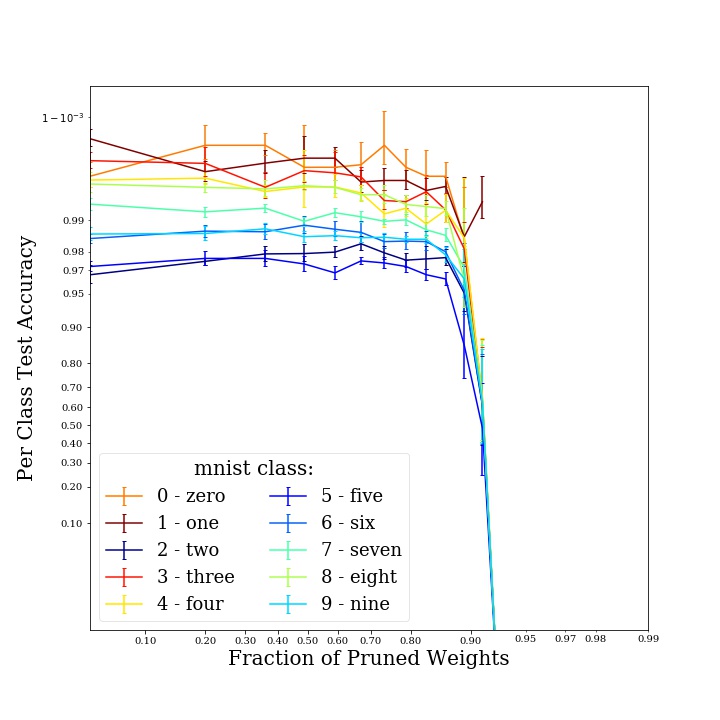}\label{fig:alexnet_mnist_perclass}}
  \hspace{1em}
  \subfloat[Total accuracy-sparsity trade-off curves for different magnitude-based structured pruning techniques.]{\includegraphics[width=0.31\textwidth]{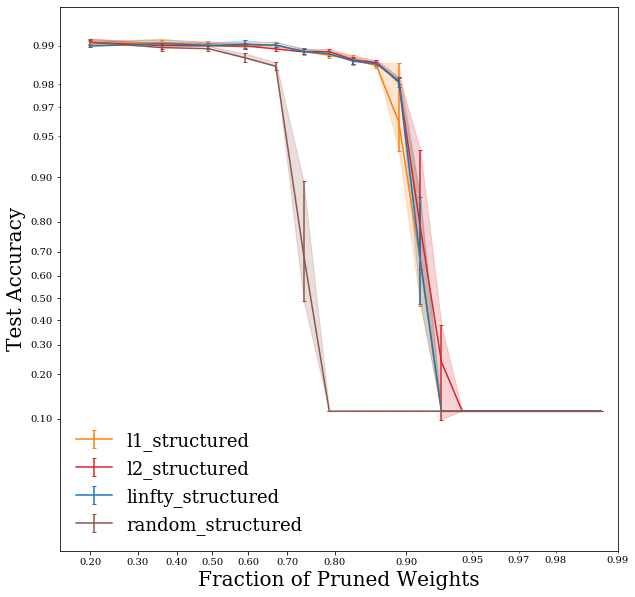}\label{fig:alexnet_mnist_accspars_struct}}
  \hspace{1em}
  \subfloat[Unfairness as a function of sparsity for different pruning techniques, where unfairness is measured distance between the highest per-class accuracy and the lowest per-class accuracy among all classes.]{\includegraphics[width=0.31\textwidth]{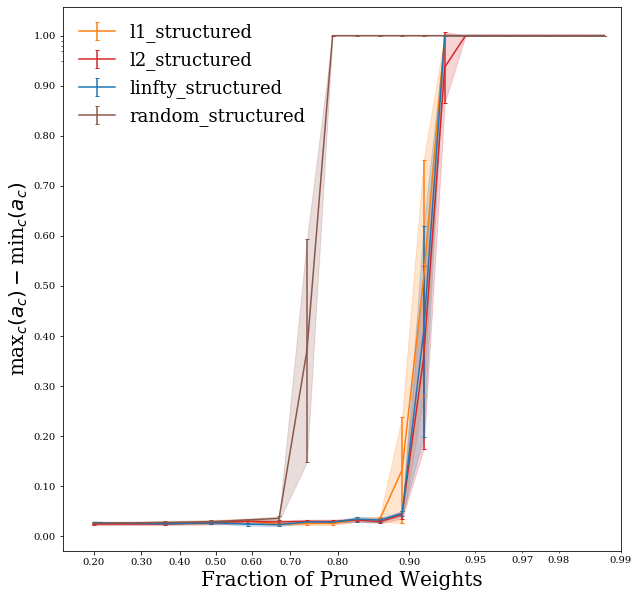}\label{fig:alexnet_mnist_fair}}
%   \vspace{0.5cm}
   \caption{AlexNet model accuracy, sparsity, and fairness curves.}\label{fig:alexnet_mnist_2d}
\end{figure}

\begin{figure}
\centering
  \subfloat[Conditioned on total accuracy $\geq 98\%$, which may correspond to a practical application scenario, networks obtained through structured sparsification are plotted in the sparsity-fairness space. Each point corresponds to the average of the accuracy and fairness results of six networks trained with seeds 0-5.
  ]{\includegraphics[width=0.45\textwidth]{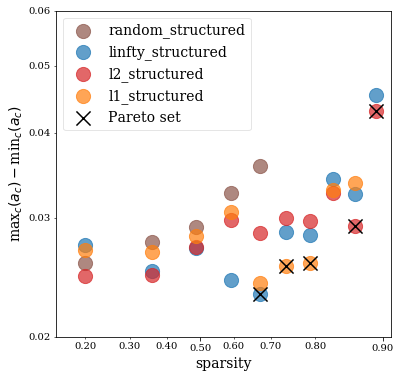}\label{fig:2d_max-min_pareto}}
  \hspace{0.5cm}
  \subfloat[Three-dimensional trade-off curves for accuracy-sparsity-fairness, where unfairness is measured as the distance between the average per-class accuracy and the lowest per-class accuracy among all classes. Points identified with black `\textsf{x}' marks are in the Pareto optimal set.]{\includegraphics[width=0.45\textwidth]{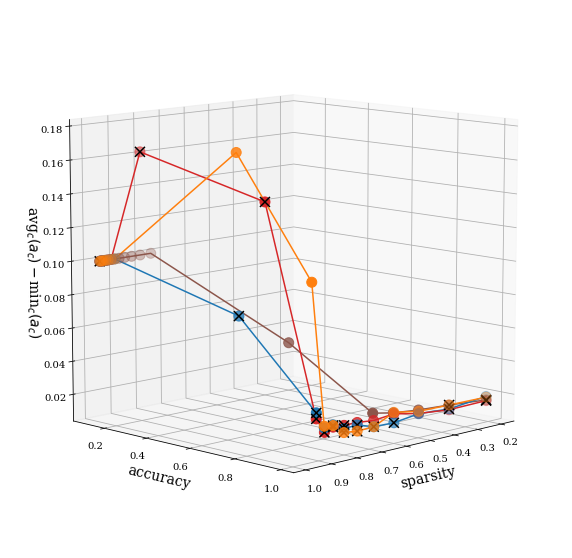}\label{fig:pareto3d}}
   \caption{Pareto frontier for pruning technique selection in a fairness-sparsity trade off scenario.}\label{fig:alexnet_mnist_paretos}
\end{figure}

Fig.~\ref{fig:2d_max-min_pareto} shows the pruned AlexNet networks that satisfy the 98\% accuracy requirement on the test set. Here, fairness -- or unfairness, rather -- is computed as the difference between the maximum and the minimum class accuracies, among the ten classes in this proxy task. The wider the gap, the more unfair the class treatment. This is just one of many valid fairness measures, but it is particularly relevant for resource allocation in systems that tend to a `winner-takes-all' state. As shown in Fig.~\ref{fig:alexnet_mnist_perclass}, this is the state high-sparsity pruned networks tend to. In general, though, the quantification of unfairness in a classifier depends on the notion of fairness one chooses, which might be application specific and context sensitive.

The networks marked with a black `\textsf{x}' in Fig.~\ref{fig:2d_max-min_pareto} lie on the Pareto frontier. All other configurations are strictly sub-optimal, and can be discarded. In other words, for any of the unmarked allocations, there exists an allocation (\textit{i.e.}, a pruned network) that is simultaneously both fairer and sparser.

It is possible, as is the case in this example, that operating points in the Pareto set correspond to different pruning techniques, thus making it impossible to immediately draw conclusions on the superiority of any pruning technique without further decisions.
However, given a value function, which encodes the relative importance of each evaluation axis in a trade-off scenario, the Pareto set can be reduced to a single operating point (or a set of equivalent operating points) that corresponds to the decision-maker's preference. For example, if, in a given context and application, fairness were valued much more (say, $100 \times$ more) than sparsity, a rational decision maker would select the blue operating point at around 68\% sparsity over any of the other allocations in Fig.~\ref{fig:2d_max-min_pareto}. Note that different stakeholders, such as model designers and subjects of the model's decisions, might have starkly different value functions with respect to the importance they assign to competing objectives, so transparency in operating point selection is key for ethical engineering and decision making.

\subsubsection{Limitations}
As we experiment with some plausible examples of fairness metrics that one might consider, we acknowledge the incompatibility of different fairness metrics, the conundrum of trying to reconcile fairness at different scales, and the fact that the few metrics considered in this work in no way make up an exhaustive list~\cite{friedler2016impossibility}. In practice, metric selection is best guided by the specific research or implementation context. While we focus on class fairness in this analysis, we invite others to explore further context-specific dimensions of evaluation. The methodology presented in this work can be naturally extended to more than three objectives of interest.

\section{Discussion}
As machine learning practitioners, it is important that we remain mindful of the consequences of our choices and interventions on models, and their effects not only on pure, overall performance metrics, but also on measures of fairness, bias, and justice. 

This work is intended to promote awareness of the extent of the variability engineers and researchers can expect to observe in the performance of pruned models across subgroups and individuals, to once again communicate limitations of standard evaluation metrics, and to inform the larger conversation around transparency and reporting of design choices.
We invite the community to be thoughtful about harmful side effects of what would otherwise appear to be uncontroversial engineering decisions, and to engineer responsibly by adopting a value-based approach to technical design.

The adoption of decision-making frameworks rooted in economics notions and traditions comes with the advantage of bridging the technical gap and language barrier between AI practitioners and stakeholders. From a policy perspective, this solution provides a single entry point for behavioral intervention, in the form of a higher abstraction layer that does not interfere with the inner workings of a pre-existing ML system in ways that may require retraining or recalibration.

\section{Conclusion}
As authors and contributors to the literature in machine learning, we felt ethically compelled to clarify intended usage modalities of neural network pruning by defining recommended ways of inspecting pruned models prior to their deployment, and evaluating pruning techniques and operating points beyond the customary sparsity-efficiency axes.
We invite the community to refrain from the characterization of novel pruning methods on classifier models exclusively in terms of the, ostensibly small, drop in total accuracy, without further investigating how that accuracy drop is absorbed by the different classes or sub-groups of interest in the task.

\section*{Broader Impact}
Our work seeks to expose shortcomings in the current pruning discourse that may have potentially harmful real-world implications for underrepresented classes and subsets of individuals subject to the decision of a pruned model.

Simply utilizing the suggested Pareto optimality framework for engineering decision-making around pruning does not guarantee the satisfaction of any fairness requirement, should not absolve from critical thinking, nor should it be misconstrued as a panacea for the lack of fairness considerations in technical decisions.

The main goal of this contribution is to foster more thoughtful design and use of pruning in ML research and systems; the lack thereof may result in the adverse consequences highlighted in this paper.

% \begin{ack}
% Use unnumbered first level headings for the acknowledgments. All acknowledgments
% More information about this disclosure can be found at: \url{https://neurips.cc/Conferences/2020/PaperInformation/FundingDisclosure}.
% Do {\bf not} include this section in the anonymized submission, only in the final paper. You can use the \texttt{ack} environment provided in the style file to autmoatically hide this section in the anonymized submission.
% \end{ack}

\bibliographystyle{plain}
\bibliography{references}

\begin{thebibliography}{10}

\bibitem{blalock2020state}
Davis Blalock, Jose Javier~Gonzalez Ortiz, Jonathan Frankle, and John Guttag.
\newblock What is the state of neural network pruning?, 2020.

\bibitem{repoqmnist}
Léon Bottou.
\newblock facebookresearch/qmnist/readme.md.
\newblock \url{https://github.com/facebookresearch/qmnist}, 2020.

\bibitem{buolamwini2018gender}
Joy Buolamwini and Timnit Gebru.
\newblock Gender shades: Intersectional accuracy disparities in commercial
  gender classification.
\newblock In {\em Conference on fairness, accountability and transparency},
  pages 77--91, 2018.

\bibitem{chen2018classifier}
Irene Chen, Fredrik~D. Johansson, and David Sontag.
\newblock Why is my classifier discriminatory?, 2018.

\bibitem{clanuwat2018deep}
Tarin Clanuwat, Mikel Bober-Irizar, Asanobu Kitamoto, Alex Lamb, Kazuaki
  Yamamoto, and David Ha.
\newblock Deep learning for classical {J}apanese literature, 2018.

\bibitem{cohen2017emnist}
Gregory Cohen, Saeed Afshar, Jonathan Tapson, and André van Schaik.
\newblock {EMNIST}: an extension of mnist to handwritten letters, 2017.

\bibitem{corbettdavies2018measure}
Sam Corbett-Davies and Sharad Goel.
\newblock The measure and mismeasure of fairness: A critical review of fair
  machine learning, 2018.

\bibitem{devries2019does}
Terrance DeVries, Ishan Misra, Changhan Wang, and Laurens van~der Maaten.
\newblock Does object recognition work for everyone?, 2019.

\bibitem{frankle2018lottery}
Jonathan Frankle and Michael Carbin.
\newblock The lottery ticket hypothesis: Finding sparse, trainable neural
  networks, 2018.

\bibitem{frankle2020early}
Jonathan Frankle, David~J. Schwab, and Ari~S. Morcos.
\newblock The early phase of neural network training, 2020.

\bibitem{friedler2016impossibility}
Sorelle~A. Friedler, Carlos Scheidegger, and Suresh Venkatasubramanian.
\newblock On the (im)possibility of fairness, 2016.

\bibitem{Gale2019-xx}
Trevor Gale, Erich Elsen, and Sara Hooker.
\newblock The state of sparsity in deep neural networks.
\newblock February 2019.

\bibitem{grother1995nist}
Patrick~J Grother.
\newblock Nist special database 19 handprinted forms and characters database.
\newblock {\em National Institute of Standards and Technology}, 1995.

\bibitem{grother2011report}
Patrick~J Grother, Patrick~J Grother, P~Jonathon Phillips, and George~W Quinn.
\newblock {\em Report on the evaluation of 2D still-image face recognition
  algorithms}.
\newblock US Department of Commerce, National Institute of Standards and
  Technology, 2011.

\bibitem{he2015deep}
Kaiming He, Xiangyu Zhang, Shaoqing Ren, and Jian Sun.
\newblock Deep residual learning for image recognition, 2015.

\bibitem{Holstein_2019}
Kenneth Holstein, Jennifer Wortman~Vaughan, Hal Daumé, Miro Dudik, and Hanna
  Wallach.
\newblock Improving fairness in machine learning systems.
\newblock {\em Proceedings of the 2019 CHI Conference on Human Factors in
  Computing Systems - CHI ’19}, 2019.

\bibitem{6327355}
B.~F. {Klare}, M.~J. {Burge}, J.~C. {Klontz}, R.~W. {Vorder Bruegge}, and A.~K.
  {Jain}.
\newblock Face recognition performance: Role of demographic information.
\newblock {\em IEEE Transactions on Information Forensics and Security},
  7(6):1789--1801, 2012.

\bibitem{krizhevsky2009learning}
Alex Krizhevsky et~al.
\newblock Learning multiple layers of features from tiny images.
\newblock 2009.

\bibitem{alexnet}
Alex Krizhevsky, Ilya Sutskever, and Geoffrey~E. Hinton.
\newblock Imagenet classification with deep convolutional neural networks.
\newblock In {\em Proceedings of the 25th International Conference on Neural
  Information Processing Systems - Volume 1}, NeurIPS 2012, page 1097–1105,
  Red Hook, NY, USA, 2012. Curran Associates Inc.

\bibitem{LeCun1990-em}
Yann LeCun, Bernhard~E Boser, John~S Denker, Donnie Henderson, R~E Howard,
  Wayne~E Hubbard, and Lawrence~D Jackel.
\newblock Handwritten digit recognition with a {Back-Propagation} network.
\newblock In D~S Touretzky, editor, {\em Advances in Neural Information
  Processing Systems 2}, pages 396--404. Morgan-Kaufmann, 1990.

\bibitem{mnist:1994}
Yann LeCun, Corinna Cortes, and Christopher J.~C. Burges.
\newblock {\em The {MNIST} database of handwritten digits.}, 1994.

\bibitem{li2016pruning}
Hao Li, Asim Kadav, Igor Durdanovic, Hanan Samet, and Hans~Peter Graf.
\newblock Pruning filters for efficient convnets, 2016.

\bibitem{model-cards}
Margaret Mitchell, Simone Wu, Andrew Zaldivar, Parker Barnes, Lucy Vasserman,
  Ben Hutchinson, Elena Spitzer, Inioluwa~Deborah Raji, and Timnit Gebru.
\newblock Model cards for model reporting.
\newblock {\em Proceedings of the Conference on Fairness, Accountability, and
  Transparency - FAT* ’19}, 2019.

\bibitem{mitchell2018predictionbased}
Shira Mitchell, Eric Potash, Solon Barocas, Alexander D'Amour, and Kristian
  Lum.
\newblock Prediction-based decisions and fairness: A catalogue of choices,
  assumptions, and definitions, 2018.

\bibitem{morcos2019ticket}
Ari~S. Morcos, Haonan Yu, Michela Paganini, and Yuandong Tian.
\newblock One ticket to win them all: generalizing lottery ticket
  initializations across datasets and optimizers, 2019.

\bibitem{netzer2011reading}
Yuval Netzer, Tao Wang, Adam Coates, Alessandro Bissacco, Bo~Wu, and Andrew~Y
  Ng.
\newblock Reading digits in natural images with unsupervised feature learning.
\newblock 2011.

\bibitem{paganini2020streamlining}
Michela Paganini and Jessica Forde.
\newblock Streamlining tensor and network pruning in {PyTorch}, 2020.

\bibitem{NEURIPS2019_9015}
Adam Paszke, Sam Gross, Francisco Massa, Adam Lerer, James Bradbury, Gregory
  Chanan, Trevor Killeen, Zeming Lin, Natalia Gimelshein, Luca Antiga, Alban
  Desmaison, Andreas Kopf, Edward Yang, Zachary DeVito, Martin Raison, Alykhan
  Tejani, Sasank Chilamkurthy, Benoit Steiner, Lu~Fang, Junjie Bai, and Soumith
  Chintala.
\newblock Pytorch: An imperative style, high-performance deep learning library.
\newblock In H.~Wallach, H.~Larochelle, A.~Beygelzimer, F.~d'Alch\'{e} Buc,
  E.~Fox, and R.~Garnett, editors, {\em Advances in Neural Information
  Processing Systems 32}, pages 8024--8035. Curran Associates, Inc., 2019.

\bibitem{phillips2003face}
P~J Phillips, Patrick~J Grother, Ross~J Micheals, D~M Blackburn, Elham Tabassi,
  and Mike Bone.
\newblock Face recognition vendor test 2002: Evaluation report.
\newblock Technical report, 2003.

\bibitem{raji2019ml}
Inioluwa~Deborah Raji and Jingying Yang.
\newblock {ABOUT ML}: Annotation and benchmarking on understanding and
  transparency of machine learning lifecycles, 2019.

\bibitem{renda2020comparing}
Alex Renda, Jonathan Frankle, and Michael Carbin.
\newblock Comparing rewinding and fine-tuning in neural network pruning, 2020.

\bibitem{vgg}
Karen Simonyan and Andrew Zisserman.
\newblock Very deep convolutional networks for large-scale image recognition,
  2014.

\bibitem{doi:10.1177/2053951717743530}
Michael Veale and Reuben Binns.
\newblock Fairer machine learning in the real world: Mitigating discrimination
  without collecting sensitive data.
\newblock {\em Big Data \& Society}, 4(2):2053951717743530, 2017.

\bibitem{verma2018fairness}
Sahil Verma and Julia Rubin.
\newblock Fairness definitions explained.
\newblock In {\em 2018 IEEE/ACM International Workshop on Software Fairness
  (FairWare)}, pages 1--7. IEEE, 2018.

\bibitem{xiang2019legal}
Alice Xiang and Inioluwa~Deborah Raji.
\newblock On the legal compatibility of fairness definitions, 2019.

\bibitem{fashionmnist}
Han Xiao, Kashif Rasul, and Roland Vollgraf.
\newblock Fashion-{MNIST}: a novel image dataset for benchmarking machine
  learning algorithms, 2017.

\bibitem{yadav2019cold}
Chhavi Yadav and Léon Bottou.
\newblock Cold case: The lost {MNIST} digits, 2019.

\bibitem{zhou2019deconstructing}
Hattie Zhou, Janice Lan, Rosanne Liu, and Jason Yosinski.
\newblock Deconstructing lottery tickets: Zeros, signs, and the supermask,
  2019.

\end{thebibliography}

\clearpage

\appendix
\section{Pruning details}
\label{appendix:pruning}
The pruning techniques explored in this work are among the ones implemented in PyTorch~\cite{NEURIPS2019_9015,paganini2020streamlining}. At all iterations, the pruning amount was set to 20\% of the unpruned weights (or units/channels, in the case of structured pruning). Structured pruning was performed along the input axis. All pruning techniques, except for global unstructured pruning, are layer-wise pruning techniques, meaning that weight (or units/channels) are compared only to other weights (or units/channels) within the same tensor.
\begin{itemize}
    \item \texttt{l1\_unstructured}: prunes individual weights in a tensor by zeroing out the one with lowest absolute value;
    \item \texttt{l1\_structured}: prunes channels or slices of a tensor by zeroing out the one with lowest absolute $L_1$ norm;
    \item \texttt{l2\_structured}: prunes channels or slices of a tensor by zeroing out the one with lowest absolute $L_2$ norm;
    \item \texttt{linfty\_structured}: prunes channels or slices of a tensor by zeroing out the one with lowest absolute $L_\infty$ norm;
    \item \texttt{random\_unstructured}: prunes individual weights in a tensor by zeroing at random;
    \item \texttt{random\_structured}: prunes channels or slices of a tensor by zeroing out channels at random;
    \item \texttt{global\_unstructured}: prunes individual weights in network by zeroing out the ones with lowest absolute value across all weights in the network;
\end{itemize}

\section{Per-class accuracy as a function of sparsity}
\begin{figure}
\centering
  \subfloat[LeNet on SVHN, random structured pruning]{\includegraphics[width=0.4\textwidth]{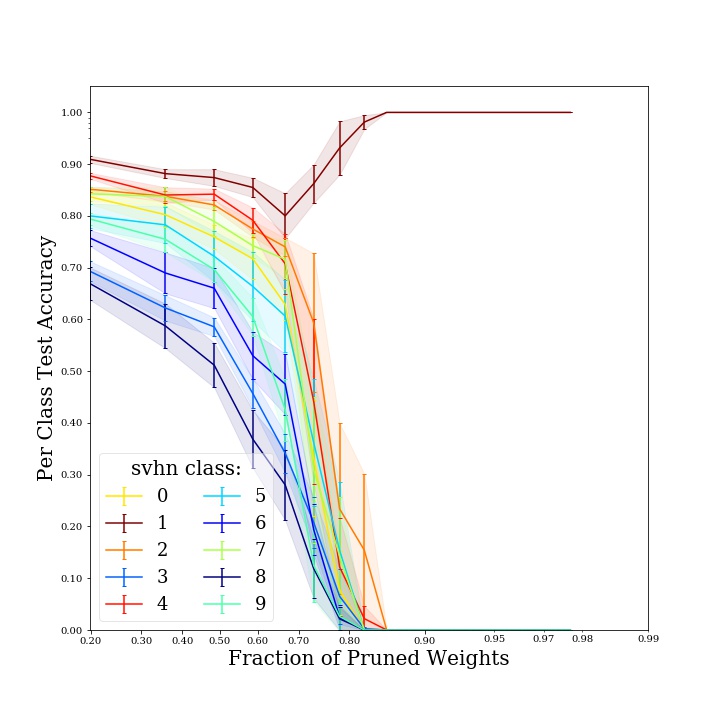}}
  \hspace{0.3cm}
  \subfloat[LeNet on FashionMNIST, $L_2$ structured pruning]{\includegraphics[width=0.4\textwidth]{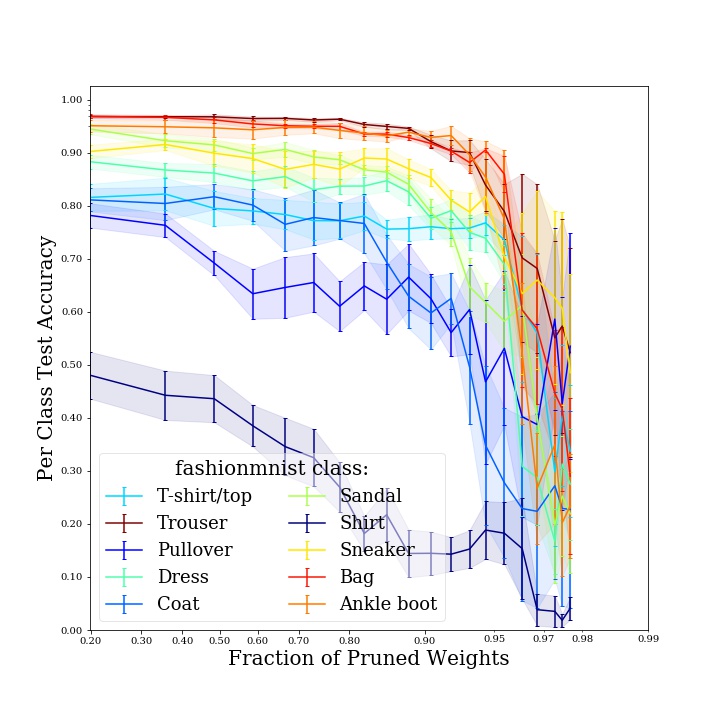}}
  \\
  \subfloat[LeNet on KMNIST, random unstructured pruning]{\includegraphics[width=0.4\textwidth]{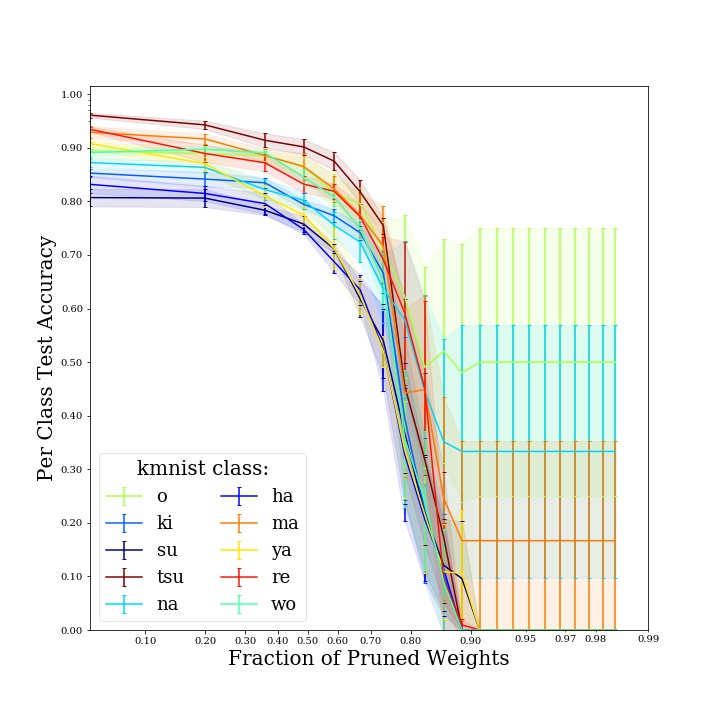}}
  \hspace{0.3cm}
  \subfloat[LeNet on CIFAR-100, global unstructured pruning]{\includegraphics[width=0.4\textwidth]{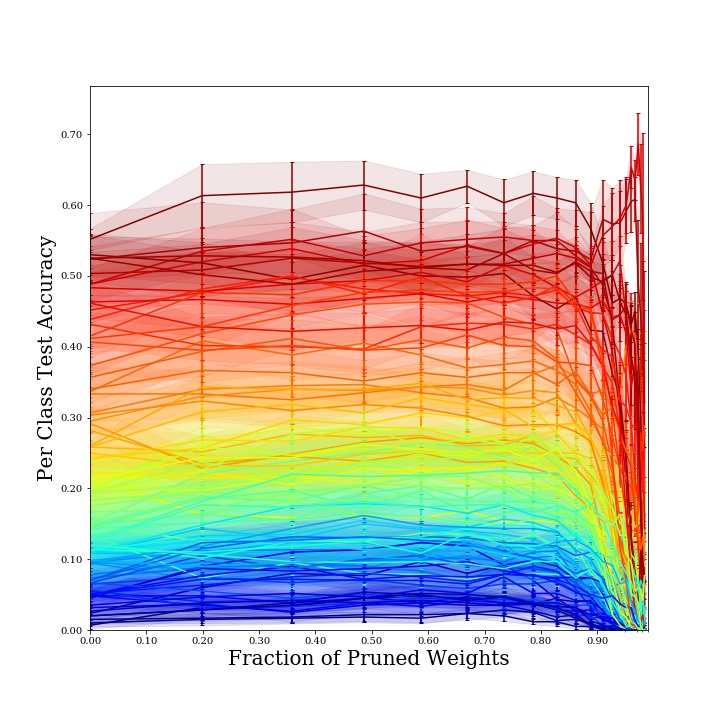}}
   \caption{Per-class accuracy-sparsity trade-off curves for a set of datasets and model architectures investigated in this work. The trade-off curve of each class is color-coded according to the accuracy the model achieves on that class prior to pruning.}\label{fig:perclass}
\end{figure}

As sparsity in a network increases as a consequence of pruning, the performance of the network eventually starts decreasing. Noticeably, however, this phenomenon happens at non-constant rate across all classes in a classification problem. Certain classes are forgotten earlier, while, on others, the network's accuracy remains high even for high levels of sparsity. The accuracy-sparsity trade-off curves in Fig.~\ref{fig:perclass} show the per-class performance degradation of a LeNet architecture trained and evaluated on the specified image classification task, and sparsified over 20 pruning iterations, with one of the pruning techniques explored in this work.

\section{QMNIST case study: further experimental details}
\label{app:qmnist}

\begin{figure}
\centering
\begin{minipage}{.45\textwidth}
    \centering
    \includegraphics[width=\textwidth]{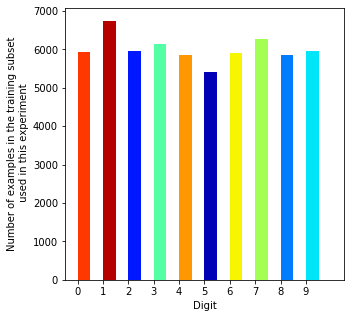}
    \caption{Histogram of examples in each class in the training subset used in these experiments. Each class's bar is color-coded according to the unpruned network's class performance, to highlight a possible dependence between the performance ranking among classes remains and class representation.}
    \label{fig:qmnist_train_distro}
\end{minipage}%
\hspace{2em}
\begin{minipage}{.45\textwidth}
  \centering
    \includegraphics[width=\textwidth]{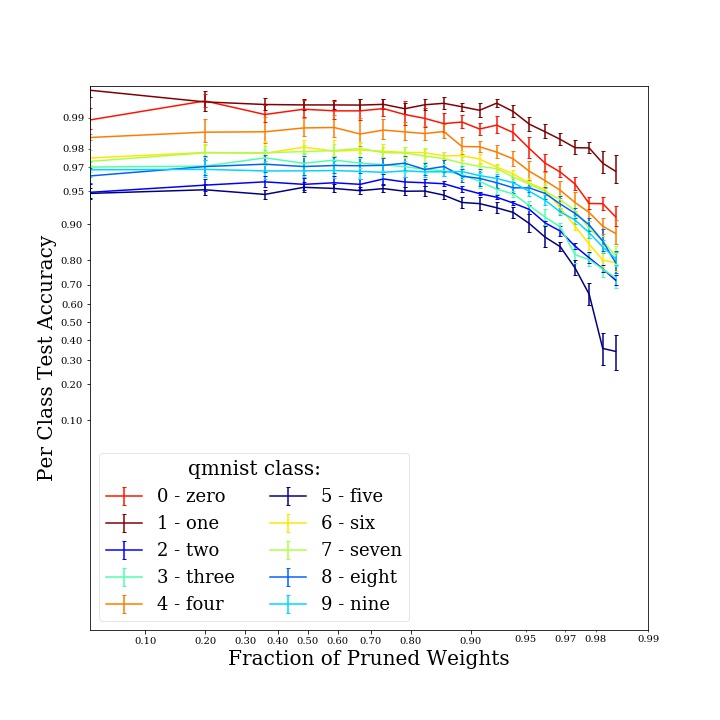}
    \caption{Per class accuracy of LeNet digit classifier on the QMNIST test set, over 20 iterations of global unstructured pruning with weight rewinding and retraining between each iteration. Each class's accuracy-sparsity curve is color-coded according to the unpruned network's class performance, to highlight the consistency with which performance ranking among classes remains almost unvaried as the network is sparsified.}
    \label{fig:qmnist_acc_perclass_GU_LeNet}
\end{minipage}
\end{figure}

QMNIST is an approximately balanced dataset, meaning that classes have a relatively comparable number of examples, without stark asymmetry. Fig.~\ref{fig:qmnist_train_distro} is the histogram of examples in each class in the subset used for training the models in this experiment. Classes are color-coded to match the legend in Fig.~\ref{fig:qmnist_acc_perclass_GU_LeNet}, which itself identifies classes by their accuracy in the unpruned model. The most over-represented class (digit 1) is the one with the highest per-class accuracy, while the most under-represented class (digit 5) is the one with the lowest per-class accuracy, suggesting that class imbalance, although minimal, may have an impact on class accuracy. Others factors also play a role, including class complexity and variability.

\begin{figure}
\centering
  \subfloat[Per-individual pruned model accuracy versus unpruned model accuracy, color-coded by average Euclidean distance of the writer's digits from the average class image.]{\includegraphics[width=0.45\textwidth]{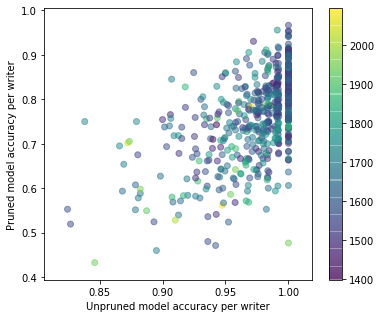}\label{fig:qmnist_avgeucl_1}}
  \hspace{0.5cm}
  \subfloat[Histogram of writers' digits' average Euclidean distance from the average class image, color-coded to match the color bar in (a).]{\includegraphics[width=0.45\textwidth]{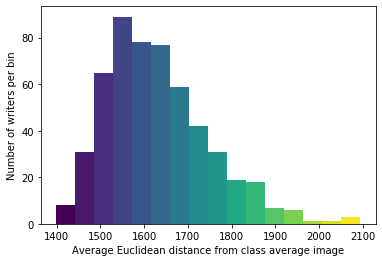}\label{fig:qmnist_avgeucl_2}}
   \caption{Impact and distribution of average Euclidean distance from the average class image for writers in the QMNIST test set.}\label{fig:qmnist_avgeucl}
\end{figure}

\begin{figure}
\centering
  \subfloat[Per-individual pruned model accuracy versus unpruned model accuracy, color-coded by mean per-image pixel activation, averaged over all digits contributed by each writer. ]{\includegraphics[width=0.45\textwidth]{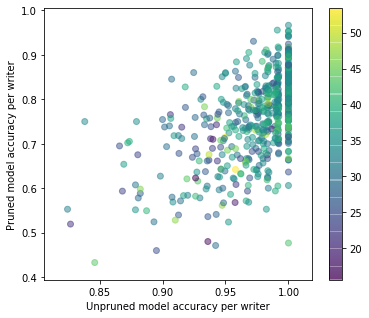}\label{fig:qmnist_avgactiv_1}}
  \hspace{0.5cm}
  \subfloat[Histogram of writers' digits' average pixel activation, color-coded to match the color bar in (a).]{\includegraphics[width=0.45\textwidth]{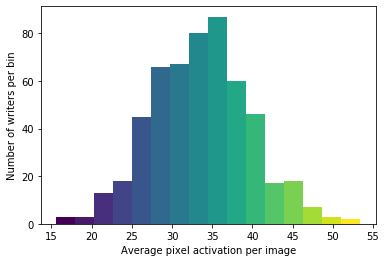}\label{fig:qmnist_avgactiv_2}}
   \caption{Impact and distribution of average image activation for writers in the QMNIST test set.}\label{fig:qmnist_avgactiv}
\end{figure}

Fig.~\ref{fig:qmnist_acc_perclass_GU_LeNet} shows how per-class accuracies drop as the network is iteratively sparsified. During this process, classes that started with a performance advantage over others end up solidifying their primacy. Noticeable is the accuracy drop underwent by the most challenging class (the digit 5) as parameters are removed from the network.

The average Euclidean distance from the class mean (itself computed on the training set) for the digits contributed by each writer is used as a measure of ``outliearness" of the writer, which we hypothesize to be a potential explanation for higher than expected drop in model performance under pruning experienced by some individuals. The plot in Fig.~\ref{fig:qmnist_acc_old_new} is represented in Fig.~\ref{fig:qmnist_avgeucl_1} with color-coding corresponding to each writer's average Euclidean distance from the mean. The distribution of this quantity is displayed alongside in Fig.~\ref{fig:qmnist_avgeucl_2}, color-coded so as to match the color bar in the corresponding scatter plot. Similar plots are shown for the average image activation per user in Fig.~\ref{fig:qmnist_avgactiv}.

\begin{figure}
    \centering
    \includegraphics[width=0.35\textwidth]{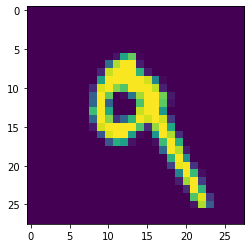}
    \hspace{1em}
    \includegraphics[width=0.35\textwidth]{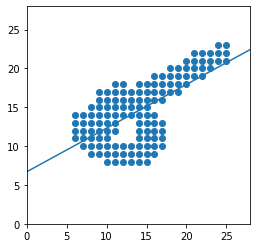}
    \caption{On the left, an image from the QMNIST test set. On the right, its binarized version with the least squares fit used to identify the image's tilt with respect to the vertical line, \textit{i.e.}, the 0 slope line, due to the 90$^\circ$ axes rotation in this representation.}
    \label{fig:tilts}
\end{figure}

\begin{figure}
    \centering
    \includegraphics[width=0.45\textwidth]{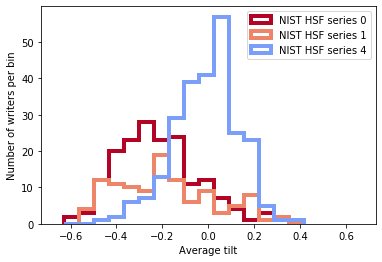}
    \hspace{1em}
    \includegraphics[width=0.45\textwidth]{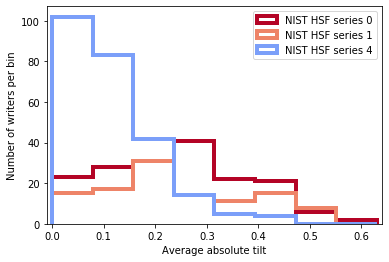}
    \caption{Distribution of average tilt and average absolute tilt per writer, separated by writer cohort.}
    \label{fig:tilts_hists}
\end{figure}

Tilt is defined as the slope of the least squares line fit through the binarized version of the image. The intercept could have been used instead, as, since the QMNIST pre-processing procedure includes a step that centers the digits by positioning their center of mass at the approximate center of the image, all linear fits should approximately pass though the center of the image, thus reducing the number of degrees of freedom. Fig.~\ref{fig:tilts} demonstrate how the slope of the line through the points that make up the digit can identify the tilt of the digit with respect to the vertical axis, which, due to the axis inversion due to the data format, corresponds to a slope of 0. Positive slopes represent tilts to the left, negative slopes represent tilts to the right. The average tilt and its absolute value are very distinctively distributed across the different writer demographic groups, as shown in the histograms in Fig.~\ref{fig:tilts_hists}. High school students, who form NIST HSF series 4, tend to draw more vertically aligned digits, on average, than NIST employees in series 0 and 1. 

\begin{figure}
\centering
  \subfloat[Per-individual pruned model accuracy versus unpruned model accuracy, color-coded by average absolute tilt of the digits produced by each writer.]{\includegraphics[width=0.45\textwidth]{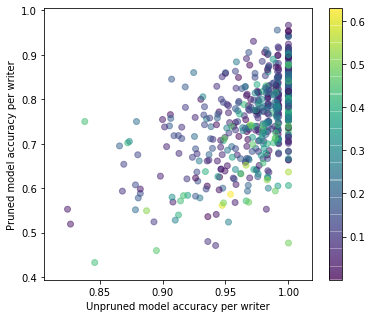}\label{fig:qmnist_avgabstilt_1}}
  \hspace{0.5cm}
  \subfloat[Histogram of average absolute tilt, color-coded to match the color bar in (a).]{\includegraphics[width=0.45\textwidth]{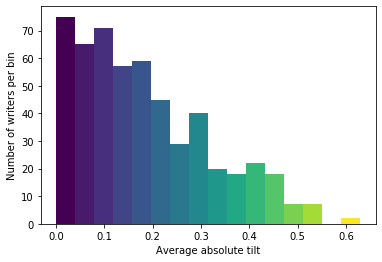}\label{fig:qmnist_avgabstilt_2}}
   \caption{Impact and distribution of average absolute tilt of digits for writers in the QMNIST test set.}\label{fig:qmnist_avgabstilt}
\end{figure}

\begin{figure}
\centering
  \subfloat[Before pruning]{\includegraphics[width=0.45\textwidth]{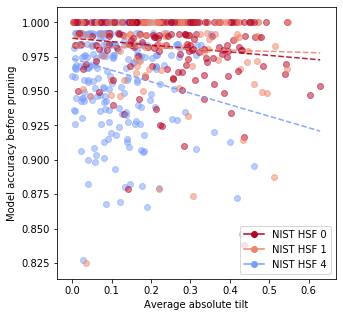}\label{fig:qmnist_avgabstilt_1}}
  \hspace{0.5cm}
  \subfloat[After pruning]{\includegraphics[width=0.45\textwidth]{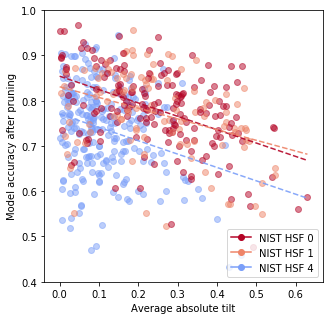}\label{fig:qmnist_avgabstilt_2}}
   \caption{Relationship between model accuracy on the digits produced by each writer and the average absolute tilt of each writer's digits.}\label{fig:before_after}
\end{figure}

The LeNet architecture utilized in this specific case study is fairly standard, consisting of two $3\times3$ Convolution-ReLU-MaxPool blocks with 6 and 16 feature maps respectively, followed by three fully-connected layers with output dimensions of 120, 84, and 10 units respectively, with ReLU used after all layers except the last.

\begin{table}[]
\renewcommand{\arraystretch}{1.2} 
\centering
\caption{Coefficients of the linear fits shown in Figs.~\ref{fig:qmnist_avgabstilt_1} and~\ref{fig:qmnist_avgabstilt_2}.}
\vspace{0.5cm}
\begin{tabular}{@{}rccccc@{}}
\toprule
     & \multicolumn{2}{c}{Before} && \multicolumn{2}{c}{After} \\ \midrule
     & intercept   & slope        && intercept   & slope       \\
     \cmidrule{2-3} \cmidrule{5-6}
\textit{hsf0} & 0.988  & -0.025  && 0.854  & -0.296 \\
\textit{hsf1} & 0.983  & -0.008  && 0.831  & -0.237 \\
\textit{hsf4} & 0.974  & -0.084  && 0.771  & -0.295 \\
full & 0.973  & 0.006   && 0.784  & -0.138 \\ \bottomrule
\end{tabular}
\label{table:tilts}
\end{table}

The effect of tilt on model the percentage change in performance is shown in Fig.~\ref{fig:tilts} in the main body of the paper; here, we measure its relationship with model performance before and after pruning, separately. First, the relationship between model accuracy on individual writers' digits, first shows in Fig.~\ref{fig:qmnist_acc_old_new}, is now presented in Fig.~\ref{fig:qmnist_avgabstilt_1} with color-coding corresponding to each writer's average absolute digit tilt. The distribution of this quantity is displayed alongside in Fig.~\ref{fig:qmnist_avgabstilt_2}, color-coded so as to match the color bar in the corresponding scatter plot.

We then plot per-writer model accuracy before and after pruning as a function of absolute tilt, and fit the distribution with a linear fit per group of writers (Fig.~\ref{fig:before_after}. While tilt already negatively affects performance even in the unpruned version of the model, especially for the cohort of writers in group \textit{hsf4}, the point is that it also affects the \textit{change} in performance after pruning, \textit{i.e.} the amount of performance degradation depends on tilt, and is not constant as a function of tilt. In other words, first, while all writers may expect some amount of performance degradation due to pruning in general, users with more tilted handwriting will experience a larger drop in accuracy from the model; second, the overall dependence of performance on tilt becomes stronger after pruning. This means that groups and individuals for whom the model was already under-performing prior to pruning are disproportionately affected by pruning, as the pruned model loses its ability to account for image tilt.

The parameters of the fits are reported in Table~\ref{table:tilts}.

\section{Fit diagnostics}
\label{appendix:fit}
\begin{figure}
    \centering
    \includegraphics[width=0.5\textwidth]{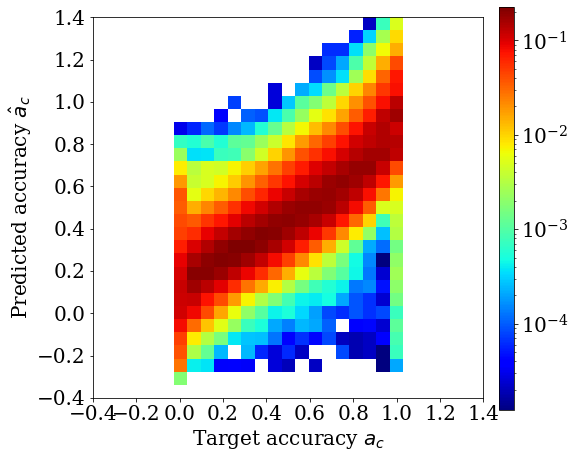}
    \caption{Relationship between the fit outcome and the target variable from data, normalized in bins of the target quantity.}
    \label{fig:pred_vs_target}
\end{figure}

\begin{figure}
\centering
  \subfloat[Histogram of residuals]{\includegraphics[width=0.47\textwidth]{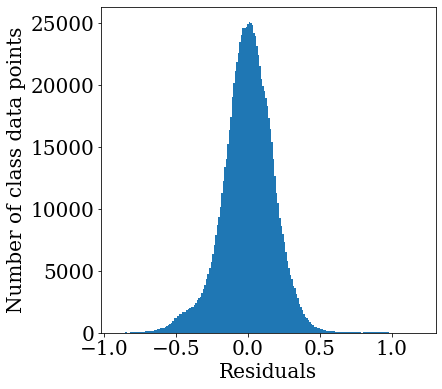}\label{fig:residuals}}
  \hspace{1em}
  \subfloat[Residuals versus target]{\includegraphics[width=0.47\textwidth]{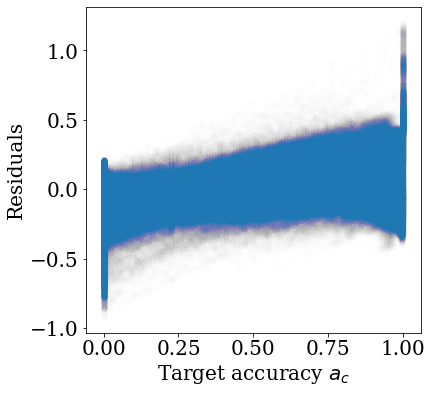}\label{fig:residuals_vs_target}}
   \caption{Distribution of residuals of the OLS fit to the class accuracies.}\label{fig:residuals_both}
\end{figure}
The Tables in this section provide further insight in the quantitative results of the OLS fit to the class accuracies across the experiments considered in this work. They can be used to gather information about the effect of each individual or cross term to the dependent variable, along with uncertainty ranges and $p$-values. 

For diagnostic purposes, we show the relationship between predicted and target class accuracy in Fig.~\ref{fig:pred_vs_target}, normalized in each bin of the target accuracy to favor regression quality interpretations over data point density reading. We use this to confirm a linear relationship between the two quantities with Gaussian-like spread, except for the critical boundary regions where the OLS assumptions fail. The histogram of fit residuals is shown in Fig.~\ref{fig:residuals}, alongside their distribution as a function of the target in Fig.~\ref{fig:residuals_vs_target}.

% \clearpage
\tiny{
\begin{center}
\begin{tabular}{lclc}
\toprule
\textbf{Dep. Variable:}                                                         &     accuracy     & \textbf{  R-squared:         } &     0.753   \\
\textbf{Model:}                                                                 &       OLS        & \textbf{  Adj. R-squared:    } &     0.753   \\
\textbf{Method:}                                                                &  Least Squares   & \textbf{  F-statistic:       } & 4.379e+04   \\
\textbf{Date:}                                                                  & Wed, 27 May 2020 & \textbf{  Prob (F-statistic):} &     0.00    \\
\textbf{Time:}                                                                  &     10:31:00     & \textbf{  Log-Likelihood:    } & 2.7551e+05  \\
\textbf{No. Observations:}                                                      &      949298      & \textbf{  AIC:               } & -5.509e+05  \\
\textbf{Df Residuals:}                                                          &      949248      & \textbf{  BIC:               } & -5.503e+05  \\
\textbf{Df Model:}                                                              &          49      & \textbf{                     } &             \\
\bottomrule
\end{tabular}
\begin{tabular}{lcccccc}
                                                                                & \textbf{coef} & \textbf{std err} & \textbf{t} & \textbf{P$> |$t$|$} & \textbf{[0.025} & \textbf{0.975]}  \\
\midrule
\textbf{Intercept}                                                              &       0.6375  &        0.010     &    62.286  &         0.000        &        0.617    &        0.658     \\
\textbf{dataset[T.cifar-100]}                                                   &       0.0074  &        0.007     &     1.010  &         0.313        &       -0.007    &        0.022     \\
\textbf{dataset[T.emnist]}                                                      &      -0.4721  &        0.010     &   -49.577  &         0.000        &       -0.491    &       -0.453     \\
\textbf{dataset[T.fashionmnist]}                                                &      -0.1293  &        0.010     &   -12.749  &         0.000        &       -0.149    &       -0.109     \\
\textbf{dataset[T.kmnist]}                                                      &      -0.1944  &        0.012     &   -15.640  &         0.000        &       -0.219    &       -0.170     \\
\textbf{dataset[T.mnist]}                                                       &      -0.0531  &        0.011     &    -4.975  &         0.000        &       -0.074    &       -0.032     \\
\textbf{dataset[T.svhn]}                                                        &       0.0342  &        0.010     &     3.339  &         0.001        &        0.014    &        0.054     \\
\textbf{model[T.LeNet]}                                                         &      -0.0651  &        0.001     &   -49.454  &         0.000        &       -0.068    &       -0.063     \\
\textbf{model[T.resnet18]}                                                      &       0.1577  &        0.001     &   186.179  &         0.000        &        0.156    &        0.159     \\
\textbf{model[T.vgg11]}                                                         &       0.0319  &        0.001     &    35.201  &         0.000        &        0.030    &        0.034     \\
\textbf{weight\_treatment[T.rewind]}                                            &       0.0670  &        0.002     &    42.561  &         0.000        &        0.064    &        0.070     \\
\textbf{pruning\_technique[T.l1\_structured]}                                   &       0.4721  &        0.005     &    93.869  &         0.000        &        0.462    &        0.482     \\
\textbf{pruning\_technique[T.l1\_unstructured]}                                 &       0.2405  &        0.005     &    46.298  &         0.000        &        0.230    &        0.251     \\
\textbf{pruning\_technique[T.l2\_structured]}                                   &       0.4781  &        0.005     &    94.198  &         0.000        &        0.468    &        0.488     \\
\textbf{pruning\_technique[T.linfty\_structured]}                               &       0.4807  &        0.005     &    92.822  &         0.000        &        0.471    &        0.491     \\
\textbf{pruning\_technique[T.random\_structured]}                               &       0.4925  &        0.005     &    95.216  &         0.000        &        0.482    &        0.503     \\
\textbf{pruning\_technique[T.random\_unstructured]}                             &       0.3984  &        0.006     &    68.576  &         0.000        &        0.387    &        0.410     \\
\textbf{weight\_treatment[T.rewind]:pruning\_technique[T.l1\_structured]}       &      -0.1592  &        0.002     &   -86.052  &         0.000        &       -0.163    &       -0.156     \\
\textbf{weight\_treatment[T.rewind]:pruning\_technique[T.l1\_unstructured]}     &      -0.0745  &        0.002     &   -38.068  &         0.000        &       -0.078    &       -0.071     \\
\textbf{weight\_treatment[T.rewind]:pruning\_technique[T.l2\_structured]}       &      -0.1578  &        0.002     &   -84.908  &         0.000        &       -0.161    &       -0.154     \\
\textbf{weight\_treatment[T.rewind]:pruning\_technique[T.linfty\_structured]}   &      -0.1834  &        0.002     &   -93.028  &         0.000        &       -0.187    &       -0.180     \\
\textbf{weight\_treatment[T.rewind]:pruning\_technique[T.random\_structured]}   &      -0.2162  &        0.002     &  -109.179  &         0.000        &       -0.220    &       -0.212     \\
\textbf{weight\_treatment[T.rewind]:pruning\_technique[T.random\_unstructured]} &      -0.2925  &        0.002     &  -133.379  &         0.000        &       -0.297    &       -0.288     \\
\textbf{np.exp(sparsity)}                                                       &      -0.5950  &        0.007     &   -85.427  &         0.000        &       -0.609    &       -0.581     \\
\textbf{np.exp(sparsity):dataset[T.cifar-100]}                                  &      -0.0424  &        0.003     &   -12.629  &         0.000        &       -0.049    &       -0.036     \\
\textbf{np.exp(sparsity):dataset[T.emnist]}                                     &       0.1515  &        0.005     &    32.138  &         0.000        &        0.142    &        0.161     \\
\textbf{np.exp(sparsity):dataset[T.fashionmnist]}                               &       0.0564  &        0.005     &    12.223  &         0.000        &        0.047    &        0.065     \\
\textbf{np.exp(sparsity):dataset[T.kmnist]}                                     &       0.0710  &        0.006     &    12.731  &         0.000        &        0.060    &        0.082     \\
\textbf{np.exp(sparsity):dataset[T.mnist]}                                      &      -0.0013  &        0.005     &    -0.264  &         0.791        &       -0.011    &        0.008     \\
\textbf{np.exp(sparsity):dataset[T.svhn]}                                       &      -0.0282  &        0.005     &    -5.975  &         0.000        &       -0.037    &       -0.019     \\
\textbf{np.exp(sparsity):pruning\_technique[T.l1\_structured]}                  &      -0.2441  &        0.002     &  -124.719  &         0.000        &       -0.248    &       -0.240     \\
\textbf{np.exp(sparsity):pruning\_technique[T.l1\_unstructured]}                &      -0.1141  &        0.002     &   -53.828  &         0.000        &       -0.118    &       -0.110     \\
\textbf{np.exp(sparsity):pruning\_technique[T.l2\_structured]}                  &      -0.2465  &        0.002     &  -125.146  &         0.000        &       -0.250    &       -0.243     \\
\textbf{np.exp(sparsity):pruning\_technique[T.linfty\_structured]}              &      -0.2422  &        0.002     &  -120.012  &         0.000        &       -0.246    &       -0.238     \\
\textbf{np.exp(sparsity):pruning\_technique[T.random\_structured]}              &      -0.2399  &        0.002     &  -117.360  &         0.000        &       -0.244    &       -0.236     \\
\textbf{np.exp(sparsity):pruning\_technique[T.random\_unstructured]}            &      -0.1536  &        0.002     &   -65.677  &         0.000        &       -0.158    &       -0.149     \\
\textbf{sparsity}                                                               &       1.3942  &        0.012     &   119.751  &         0.000        &        1.371    &        1.417     \\
\textbf{accuracy0}                                                              &       1.1384  &        0.005     &   207.671  &         0.000        &        1.128    &        1.149     \\
\textbf{accuracy0:weight\_treatment[T.rewind]}                                  &      -0.1827  &        0.002     &   -90.347  &         0.000        &       -0.187    &       -0.179     \\
\textbf{accuracy0:pruning\_technique[T.l1\_structured]}                         &      -0.0826  &        0.003     &   -25.703  &         0.000        &       -0.089    &       -0.076     \\
\textbf{accuracy0:pruning\_technique[T.l1\_unstructured]}                       &      -0.0178  &        0.003     &    -6.100  &         0.000        &       -0.024    &       -0.012     \\
\textbf{accuracy0:pruning\_technique[T.l2\_structured]}                         &      -0.0842  &        0.003     &   -25.998  &         0.000        &       -0.091    &       -0.078     \\
\textbf{accuracy0:pruning\_technique[T.linfty\_structured]}                     &      -0.0967  &        0.003     &   -28.895  &         0.000        &       -0.103    &       -0.090     \\
\textbf{accuracy0:pruning\_technique[T.random\_structured]}                     &      -0.1374  &        0.003     &   -39.518  &         0.000        &       -0.144    &       -0.131     \\
\textbf{accuracy0:pruning\_technique[T.random\_unstructured]}                   &      -0.1687  &        0.004     &   -45.827  &         0.000        &       -0.176    &       -0.161     \\
\textbf{accuracy0\_quartile}                                                    &      -0.0098  &        0.001     &   -15.041  &         0.000        &       -0.011    &       -0.009     \\
\textbf{imbalance}                                                              &       1.2429  &        0.023     &    53.527  &         0.000        &        1.197    &        1.288     \\
\textbf{class\_entropy}                                                         &      -0.0237  &        0.001     &   -43.313  &         0.000        &       -0.025    &       -0.023     \\
\textbf{np.exp(sparsity):accuracy0}                                             &      -0.1907  &        0.002     &   -91.104  &         0.000        &       -0.195    &       -0.187     \\
\textbf{accuracy0:accuracy0\_quartile}                                          &       0.0261  &        0.001     &    38.494  &         0.000        &        0.025    &        0.027     \\
\bottomrule
\end{tabular}
\begin{tabular}{lclc}
\textbf{Omnibus:}       & 38897.417 & \textbf{  Durbin-Watson:     } &     0.914   \\
\textbf{Prob(Omnibus):} &    0.000  & \textbf{  Jarque-Bera (JB):  } & 110954.176  \\
\textbf{Skew:}          &   -0.159  & \textbf{  Prob(JB):          } &      0.00   \\
\textbf{Kurtosis:}      &    4.644  & \textbf{  Cond. No.          } &      624.   \\
\bottomrule
\end{tabular}
%\caption{OLS Regression Results}
\end{center}

Warnings: \newline
 [1] Standard Errors are heteroscedasticity and autocorrelation robust (HAC) using 3 lags and with small sample correction
}

\normalsize{

\section{Datasets}

This work builds on publicly available, standard datasets for reproducibility purposes. For all datasets, the training set is partitioned into a 20\% random split for validation and an 80\% split for training. The split is controlled by the experimental seed for reproducibility. Datasets are pulled and verified using \texttt{torchvision}. All images per dataset are normalized and, at train time, are augmented via a random crop and a horizontal flip. All images are ultimately resized to fit the input layer of the networks employed in this work. The dataset sizes and image sizes listed below represent the original dataset formats prior to our augmentation procedure. The code attached the supplemental material will shed light on the specific implementation details of the data loading process.

\subsection{MNIST, KMNIST, and Fashion-MNIST}

MNIST~\cite{mnist:1994}, Kuzushiji-MNIST (or KMNIST)~\cite{clanuwat2018deep}, and Fashion-MNIST~\cite{fashionmnist} each consists of 60,000 training and 10,000 test grey-scale images. 

\subsection{QMNIST}

QMNIST~\cite{repoqmnist,yadav2019cold} consists of the standard 60,000 training images from MNIST, but reconstructs a total of 60,000 test images with attributes, where the 10,000 test samples are included in this reconstruction procedure (see Sec.~\ref{app:qmnist} for more details).

\subsection{EMNIST}

The \emph{ByClass} EMNIST~\cite{cohen2017emnist} split consists of 814,255 examples (697,932 training, and 116,323 validation) drawn from 62 unbalanced classes. These are grey-scale images that represent digits and letters in the Latin alphabet.

\subsection{CIFAR-10 and CIFAR-100}

CIFAR-10 and CIFAR-100 consist of 60,000 examples -- 50,000 training images, and 10,000 test images, all of which are $32\times32$ color images~\cite{krizhevsky2009learning}. CIFAR-10 consists of 10 classes with 6,000 per class, and CIFAR-100 consists of 100 classes (fine labels), with 600 examples per class. These are datasets of natural images.

\subsection{SVHN}

SVHN~\cite{netzer2011reading} (Format 2 - cropped digits) consists of 73,257 training and 26,032 test examples. All examples are color images of housing numbers obtained from Street View.

}
\end{document}